\pdfoutput=1

\documentclass[11pt,table]{article}
\usepackage[most]{tcolorbox}
\usepackage{graphicx}
\usepackage{booktabs} 
\usepackage{acl}
\usepackage{booktabs}
\usepackage{graphicx} 
\usepackage{times}
\usepackage{latexsym}

\usepackage[T1]{fontenc}
\usepackage{amssymb}
\usepackage[utf8]{inputenc}

\usepackage{microtype}

\usepackage{inconsolata}
\usepackage{amsmath}
\usepackage{graphicx}
\usepackage{hyperref}
\title{Beyond the Textual: Generating Coherent Visual Options for MCQs}

\author{
 \textbf{Wanqiang Wang\textsuperscript{1}},\quad
 \textbf{Longzhu He\textsuperscript{2}\thanks{Corresponding author.},\quad
 \textbf{Wei Zheng\textsuperscript{1}}} \\
 \\
 \textsuperscript{1}Beijing Normal University, Beijing, China\\
 \textsuperscript{2}Beijing University of Posts and Telecommunications, Beijing, China\\
 \texttt{\{wwq2001, 05166\}@mail.bnu.edu.cn, helongzhu@bupt.edu.cn}
}

\begin{document}
\definecolor{myblue}{RGB}{22,93,225}
\maketitle
\begin{abstract}
Multiple-choice questions (MCQs) play a crucial role in fostering deep thinking and knowledge integration in education. However, previous research has primarily focused on generating MCQs with textual options, but it largely overlooks the visual options. Moreover, generating high-quality distractors remains a major challenge due to the high cost and limited scalability of manual authoring. To tackle these problems, we propose a Cross-modal Options Synthesis (\texttt{\textbf{CmOS}}), a novel framework for generating educational MCQs with visual options. Our framework integrates Multimodal Chain-of-Thought (MCoT) reasoning process and Retrieval-Augmented Generation (RAG) to produce semantically plausible and visually similar answer and distractors. It also includes a discrimination module to identify content suitable for visual options. Experimental results on test tasks demonstrate the superiority of \texttt{\textbf{CmOS}} in content discrimination, question generation and visual option generation over existing methods across various subjects and educational levels.

\end{abstract}

\section{Introduction}

In the field of education, multiple-choice questions (MCQs) play a crucial role in promoting deep thinking and knowledge integration. Prior studies have shown that  well-written MCQs can support learner engagement in higher levels of cognitive reasoning such as application or synthesis of knowledge \citep{davis2009tools,zaidi2018pushing}. Among the components of MCQs, the difficulty and relevance of distractors are key indicators of question quality \citep{gierl2017developing, kumar2023novel}. However, systematically crafting quality assessment questions and distractors in education is a crucial yet time-consuming, expertise-driven undertaking that calls for innovative solutions \citep{indran2024twelve}.
\begin{figure}[t]
  \includegraphics[width=\columnwidth] {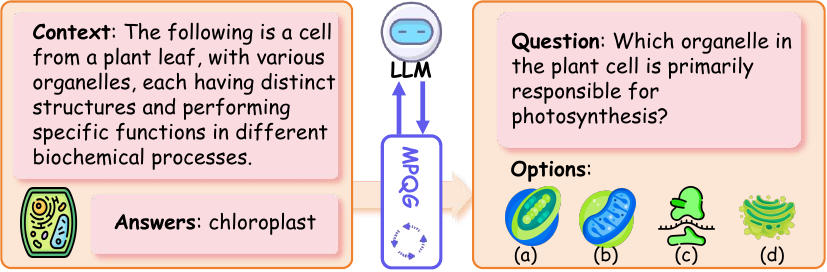}
  \caption{An example of the \texttt{\textbf{CmOS}} Framework, showing how it generates a visual MCQ to identify the part responsible for photosynthesis, using content that includes an image, an answer, and the context about plant cell.}
  \label{fig1}
\end{figure}

With the rapid advancement of large language models (LLMs) that demonstrate remarkable capabilities in scientific question answering, researchers have begun to explore their potential in the automatic generation of educational questions to alleviate human labor, including objective and open-ended formats \cite{cao2021controllable,rodriguez2022end}. Prior to this study, research on MCQs generation using LLMs primarily focused on two areas: generating questions and textual options based on textual input \citep{cao2021controllable,le2022unsupervised}, and incorporating multimodal inputs to generate questions and textual options enriched with visual information \citep{yeh2022multi,wang2023multiqg,luo2024chain}. However, the task of generating MCQs with visual options remains largely unexplored. According to Mayer’s cognitive theory of multimedia learning \citep{mayer2005cognitive}, visual stimuli play an indispensable role in promoting learners’ cognitive engagement by facilitating dual-channel processing, reducing extraneous cognitive load, and enhancing the integration of new information with prior knowledge.

There are three major challenges in generating MCQs with visual options. Firstly, not all MCQ options are suitable for visual representation \citep{butler2018multiple}. For example, mathematical computation problems typically rely on precise numerical or symbolic expressions which is difficult to convey through static images. Secondly, MCQs with visual option require explicit scaffolding for visual analysis; otherwise, students may incur unnecessary cognitive overload \citep{kim2011scaffolding}. Lastly, current Text-to-Image (T2I) models face key challenges in visual option generation: (\texttt{i}) The inherent variability in visual option generation often leads to inaccurate or unrealistic outputs. (\texttt{ii}) The educational domain lacks specialized image repositories to meet the precise visual needs of academic content. As shown in Figure \ref{fig1}, the goal of our work is to generate an appropriate question and visual options from an input consisting of an image, context, and the answer, via processing pipeline.

To address these challenges, we propose a novel framework named \textbf{C}ross-\textbf{m}odal \textbf{O}ptions \textbf{S}ynthesis (\texttt{\textbf{CmOS}}), integrating Multimodal Chain-of-Thought (MCoT) reasoning with Retrieval-Augmented Generation (RAG) to generate MCQs with visual options. Specifically, the framework employs an Multimodal Large Language Model (MLLM) to encode multimodal content and embeds it into a four-stage MCoT architecture that separates content discrimination, question and reason generation, alternative pairs screening, and visual options generation. To improve the quality of visual options, we leverage RAG to retrieve similar images from an external educational image database as templates for generation, and then the MLLM and the T2I model are required to optimize based on the templates.

The practical and impactful contributions of this work in MCQs can be summarized as follows:
\begin{itemize}
    \item We first explore how to generate high-quality MCQs with visual options, addressing a key gap in current MCQs generation research.

    \item We propose a framework named \textbf{\texttt{CmOS}}, which generates questions through multiple questions screening, and produces plausible visual options by providing templates and tuning.
    \item Experimental results on three tasks indicate that \texttt{\textbf{CmOS}} achieves superior performance over existing methods, effectively harnessing the capabilities of both MLLMs and T2I models.
\end{itemize}

\section{Related Works}
\subsection{Automatic Question Generation}


Automatic Question Generation (AQG) is a technology that addresses the high costs and inefficiencies of manually creating educational questions \citep{brown2005automatic}. Early studies in AQG primarily focus on textual question generation, comprising two categories of generation methods. The first is the Level of Understanding approach, encompassing syntax-based \citep{afzal2011unsupervised,afzal2014automatic} and semantic-based methods \citep{ai2015semi,afzal2014automatic}. The second is the Procedure of Transformation approach, including template-based \citep{kusuma2018automatic,kusuma2022automatic}, rule-based \citep{singhal2014automated,singhal2015framework}, and statistical methods \citep{kumar2015automatic,uto2023difficulty}. However, these methods rely heavily on text as both input and output, exhibiting limited capability in processing multimodal information. In terms of question types, prior research has mainly addressed open-ended and MCQs \citep{kurdi2020systematic,mulla2023automatic}, with the latter receiving increasing attention due to their standardized format and ease of automated assessment \citep{touissi2022does,al2024automatic,newton2024chatgpt}. With the swift progress of LLMs and MLLMs, researchers have begun to explore their application in generating high-quality educational questions \citep{mulla2023automatic,yadav2023contextualizing,newton2024chatgpt,al2024automatic}. Prompt engineering and fine-tuning have been employed to enhance the effectiveness of these models in question generation, with growing interest in their potential for multimodal MCQs generation \citep{zhao2022educational,wang2023multiqg,luo2024chain}. 
However, current studies on multimodal MCQs generation predominantly focus on aligning images with question context, while significantly overlooking the integration of visual information into option generation progress.

\subsection{Chain-of-Thought Prompt}
Chain-of-Thought (CoT) prompt is an important technique aimed at enhancing the multi-step reasoning capabilities of LLMs. Initial work by \citet{wei2022chain} showed that few-shot CoT could significantly improve performance on arithmetic and commonsense tasks. Later, researchers introduced zero-shot variants, such as appending simple phrases like ``Let’s think step by step''. This approach activated reasoning in LLMs without requiring additional examples \citep{kojima2022large}. To minimize the need for manually crafted demonstrations, methods like Auto-CoT were developed  \citep{zhang2022automatic}. These ways leverage LLMs to generate reasoning examples with minimal supervision. Further reliability improvements came from self-consistency decoding, which samples multiple reasoning paths and selects the most frequent answer \citep{wang2022self}. Beyond text-based tasks, researchers have extended CoT to multimodal content and proposed a method named Multimodal Chain-of-Thought (MCoT) \citep{zhang2023multimodal}. MCoT integrates textual and visual reasoning through several implementation strategies. These include two-stage pipelines that separate rationale generation from answer prediction \citep{zhang2023multimodal}, dual-guidance approaches that disentangle visual and textual reasoning \citep{jia2024dcot}, and methods that interleave image regions with textual steps \citep{mitra2024compositional, hu2024visual}. These designs enable MLLMs to perform well across various multimodal benchmarks. More related work is discussed in Appendix ~\ref{rw}.

\section{Method}
\begin{figure*}[t]
  \includegraphics[width=\linewidth] {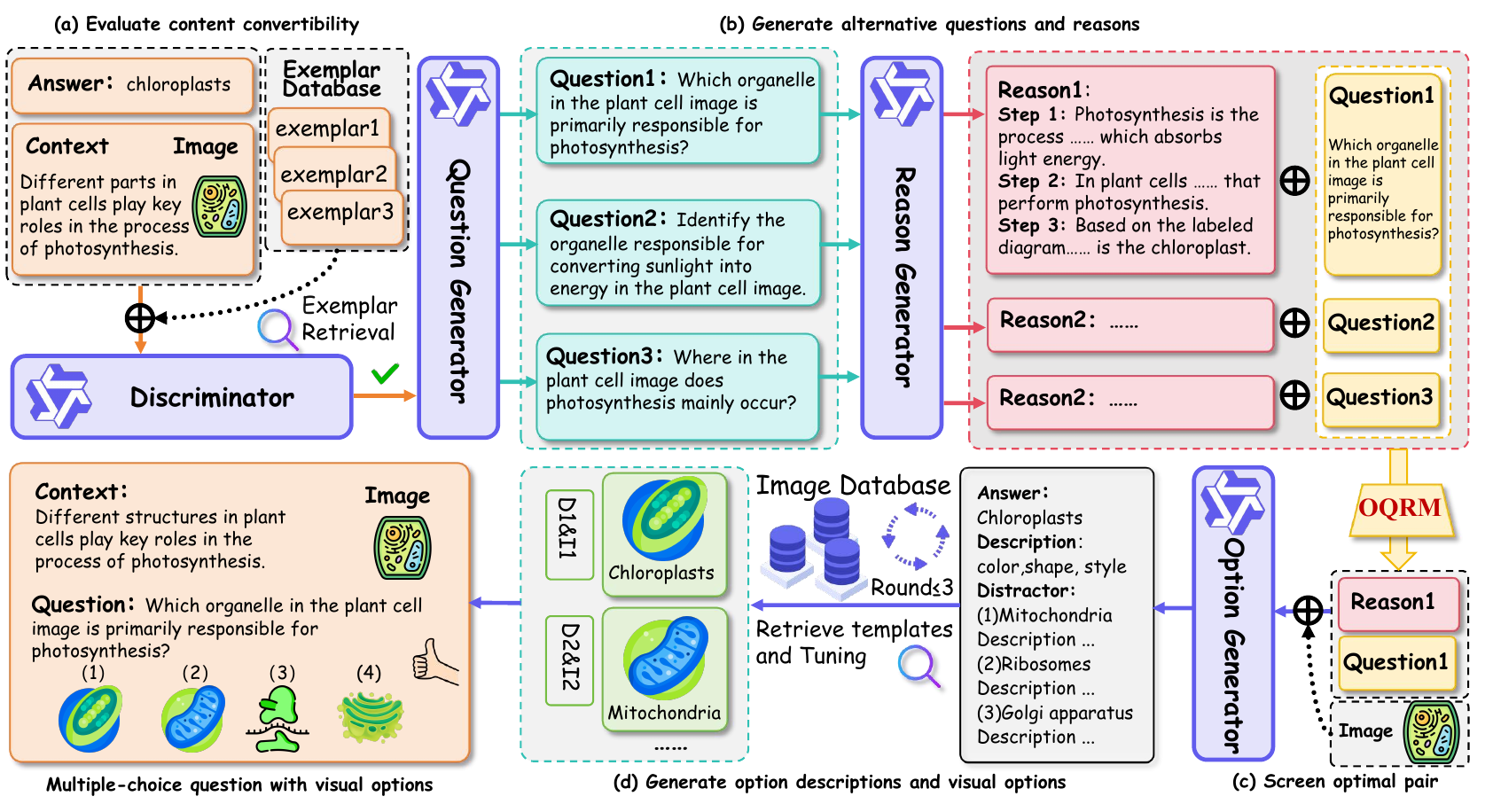}
  \caption{Overview of multimodal educational questions and visual options generation: (a) Evaluate content convertibility: we concatenate the best retrieved exemplar with the instances for content discrimination; (b) Generate alternative questions and reasons: we use prompts to require MLLM to generate diverse questions and reasons; (c) Screen optimal pair: we select the optimal question-reason pair based on internal and external consistency; and (d) Generate option descriptions and visual options: we generate visual options using templates and tuning methods.}
  \label{framewrok}
\end{figure*}
\textbf{Problem Definition} Given an input consisting of $\mathcal{C} = (T, I, A)$, where \( T \) represents the text and \( I \) represents the image, and \( A \) denotes the answer, the task is to generate an output that includes a high-quality question \( Q \), a visual option \( A' \) corresponding to the answer, and multiple visual distractors \( D_s \), each associated with a textual distractor.

\subsection{Model Architecture}
As illustrated in Figure \ref{framewrok}, our \texttt{\textbf{CmOS}} consists of four distinct stages: (\texttt{i}) evaluating content convertibility, (\texttt{ii}) generating alternative questions and reasons, (\texttt{iii}) selecting the optimal pair, and (\texttt{iv}) generating option descriptions and visual options. To address the complexity of content discrimination, question and reason generation, and visual option generation, we introduce the MCoT prompt. This method enables MLLMs to identify suitable content for conversion, generate questions and reasons towards the answer, and guide the generation of visual options, based on dynamically exemplars. Specifically, in stage (\texttt{a}), we retrieve relevant multimodal exemplars from an external repository, which includes the original content (text and image) along with convertibility judgments and reasons. They are used to construct dynamic MCoT prompts, enhancing the model's accuracy in content discrimination. In stage (\texttt{b}), we provide three reference processes for each question-reason pair, directing the MLLM to emulate the exemplar reasons. In stage (\texttt{c}), the Optimal Question-Reason Matching (OQRM) module selects the optimal question-reason pair based on internal and external consistency. In stage (\texttt{d}), the option generator produces textual options and their corresponding visual descriptions. Based on these descriptions, relevant images are dynamically retrieved from an external image database, serving as reference templates for generating visual options. Iterative evaluation and tuning using a Text-to-Image (T2I) model further improve the quality of the visual options.

\subsection{Exemplars Construction and Retrieval}
We construct exemplars using Qwen2.5-VL-72B \citep{bai2023qwen}, a MLLM with strong visual understanding and instruction-following capabilities. The exemplars include two parts: (i) the original MCQ's context, image, and answer; and (ii) the judgment and reason about its convertibility. We extract 482 questions from the ScienceQA test dataset as the foundation for exemplars construction.

After constructing the exemplars dataset $\mathcal{D_E}$, we introduce an exemplar retrieval mechanism to assist the discriminator in accurately and efficiently determining the convertibility of the given content. Specifically, we adopt the latest FARE encoder \citep{schlarmann2024robust}, which enhances robustness over CLIP. The FARE consists of a visual encoder $\phi: I \to \mathbb{R}^D$ and a text encoder $\psi: T \to \mathbb{R}^D$. For a given instance $S$ to be converted, we separately encode its corresponding context $t$, answer $a$, and image $i$ as $\mathcal{V}_t$, $\mathcal{V}_a$, and $\mathcal{V}_i$. Let $\mathcal{I}=\{1,2,\ldots,N\}$ be the index set of exemplars. 
Denote the encoded vectors of the $j$-th exemplar's text, answer, and image as 
$\mathcal{V}_t^j$, $\mathcal{V}_a^j$, and $\mathcal{V}_i^j$, respectively. We compute cross-modality similarities:
\begin{equation}
\mathrm{Sim}_m^j=\frac{\mathcal{V}_m^j \cdot \mathcal{V}_m}{\|\mathcal{V}_m^j\|_2\,\|\mathcal{V}_m\|_2},
\quad m\in\{t,a,i\}.
\end{equation}
Select the exemplar by the maximum similarity:
\begin{equation}
j^\star=\underset{j\in\mathcal{I}}{\arg\max}\ \max\big(\mathrm{Sim}_t^j,\mathrm{Sim}_a^j,\mathrm{Sim}_i^j\big).
\end{equation}
The exemplar with index $j^\star$ is concatenated with instance $S$ before being fed into the discriminator.


\subsection{Optimal Question-Reason Match}

After being processed by the question generator and reason generator with MCoT, several question-reason pairs are produced. To determine which pair is most suitable for generating multiple visual options, we introduce the Optimal Question-Reason Match module (OQRM), which can calculates a Total Match Score (TMS) for each question-reason pair to effectively identify the optimal candidate.

The visual encoder $\phi$ encodes the image \textit{i} from the instance $S$ to obtain a high-dimensional vector representation $\mathcal{V}_i$. The text encoder $\psi$ processes three textual components: the newly generated question \textit{q}, the reason \textit{r}, and the phrase \texttt{"a photo of A"} \textit{p}, yielding vectors $\mathcal{V}q$, $\mathcal{V}r$, and $\mathcal{V}p$, respectively. The TMS for the $k$-th ($k \in {1, 2, \ldots, m}$) question-reason pair $(q_k, r_k)$ is calculated as the weighted sum of two similarity scores: internal consistency $\mathcal{C}{int_k}$ and external consistency $\mathcal{C}{ext_k}$. $\mathcal{C}{int_k}$ measures the coherence of the question-reason pair within its embedding space $\mathbb{R}^{D}$, identifying the pair closest to the center $(\mathcal{C}_Q, \mathcal{C}_R)$ to ensure coherence. $\mathcal{C}{ext_k}$ evaluates the similarity between the pair and the original content. This approach aligns with the self-consistency method proposed by \citep{wang2022self}, which selects the most consistent reason path to mitigate hallucinations and avoid generating irrelevant distractors.

\begin{equation}
\begin{aligned}
\mathcal{C}_{int_k}= \frac{\mathcal{V}_{q_k}\cdot\mathcal{C}_Q}{\|\mathcal{V}_{q_k}\|_2\|\mathcal{C}_Q\|_2}+\frac{\mathcal{V}_{r_k}\cdot\mathcal{C}_R}{\|\mathcal{V}_{r_k}\|_2\|\mathcal{C}_R\|_2}
\end{aligned}
\end{equation}
where $\mathcal{C}_Q$=$\frac{1}{m} \sum_{k=1}^m\mathcal{V}{q_k}$, $\mathcal{C}_R$=$\frac{1}{m} \sum_{k=1}^m\mathcal{V}{r_k}$.

\begin{equation}
\begin{aligned}
\mathcal{C}_{ext_k}= \frac{\mathcal{V}_i\cdot\mathcal{V}_{r_k}}{\|\mathcal{V}_i\|_2\|\mathcal{V}_{r_k}\|_2}+\frac{\mathcal{V}_{q_k}\cdot\mathcal{V}_p}{\|\mathcal{V}_{q_k}\|_2\|\mathcal{V}_p\|_2}
\end{aligned}
\end{equation}
Finally, considering the hyperparameter \(\alpha\) to optimally balance $\mathcal{C}{int_k}$ and $\mathcal{C}{ext_k}$, we 
 select the question-reason pair $(q^*, r^*)$ with the highest TMS:
\begin{equation}
(q^*, r^*) = \arg\max_{(q_k, r_k)} \sum \left( \alpha \mathcal{C}_{int_k} + \mathcal{C}_{ext_k} \right)
\end{equation}

\subsection{Visual Options Generation}
We require the Option Generator to produce \(t\) options and their visual descriptive information (including a correct option and  \(t-1\) distractor options) based on the optimal question-reason pair. Based on the RAG, we propose an adaptive method to generate visual options. This method integrates the excellent text-image alignment ability of the MLLM $\mathcal{G}$ with the generation and enhancement capabilities of the Text-to-Image (T2I) model $\mathcal{P}$.

We construct an image database $\mathcal{D}$ by collecting images $i_j$ from the ScienceQA and generating corresponding captions $c_j$ ($j = 1, 2, \cdots, s$) using MLLM. Given an option description $d_i$ ($i = 1, 2, \cdots, t$), we compute the total similarity between $d_i$ and each image-caption pair as following:
\begin{equation}
\small
Sim_{ij} = 
\underbrace{\beta\frac{\mathcal\phi(i_j) \cdot \mathcal\psi(d_i)}{\|\mathcal\phi(i_j)\|_2 \|\mathcal\psi(d_i)\|_2}}_{\text{image-description}}+ 
\underbrace{\frac{\mathcal\psi(c_j) \cdot \mathcal\psi(d_i)}{\|\mathcal\psi(c_j)\|_2 \|\mathcal\psi(d_i)\|_2}}_{\text{caption-description}}
\end{equation}

For each option description \( d_i \), we retrieve the image \( i_j \) from \(\mathcal{D}\) with the highest \( Sim^{\text{total}}_{ij} \) as template. Using description \( d_i \) and template\( i_j \), the image generator \(\mathcal{P}\) produces a visual image, which is then evaluated by \(\mathcal{G}\) to obtain a similarity score \( Sim_k \). If \( Sim_k \) meets or exceeds the threshold \(\sigma=0.8\), the image is accepted. Otherwise, \(\mathcal{G}\) provides suggestions \( S \) to \(\mathcal{P}\) for iterative tuning of the generated image up to three rounds.

\begin{figure*}[t]
  \includegraphics[width=\linewidth] {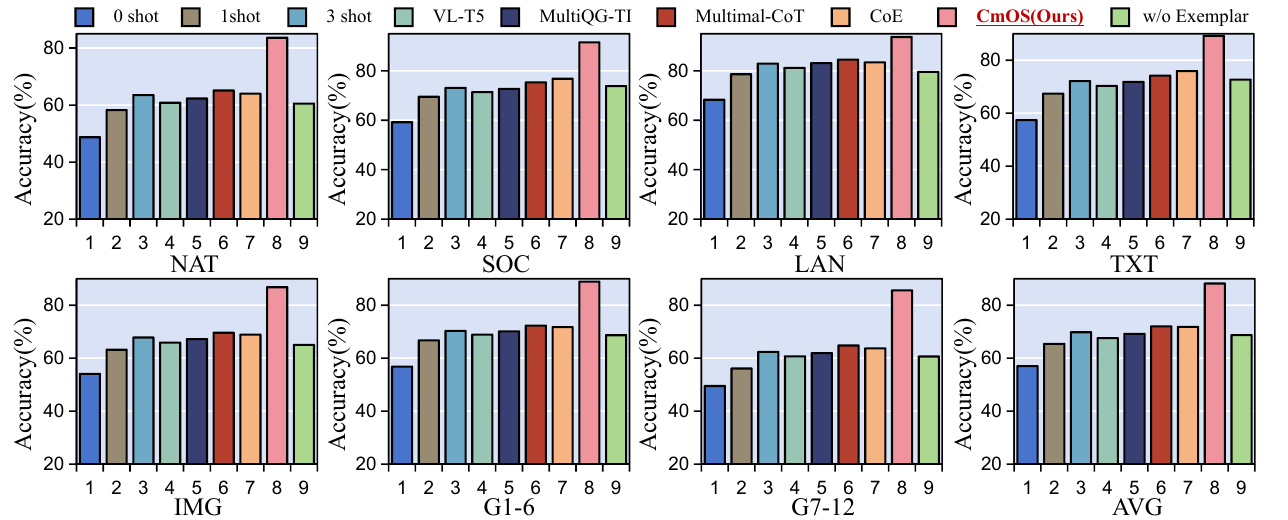}
  \caption{Automatic evaluation results of content discrimination. In terms of accuracy, regardless of subject, modality, or grade, our framework outperforms all baselines, with an average accuracy of \textbf{88.2}.}
  \label{cd}
\end{figure*}
\section{Experiment}
\subsection{Experimental Setups}
\paragraph{Dataset} To evaluate our framework's performance in content discrimination, question and visual option generation, we constructed our test datasets based on the ScienceQA benchmark test set \(\mathcal{D}_O\)\citep{lu2022learn}. Each record in $\mathcal{D}_O$ contains the context, question, options, answer, grade, subject, and so on. We randomly sampled $482$ instances from $\mathcal{D}_O$ and added ``convertible'' and ``reason'' to form the exemplar set $\mathcal{D}_E$. In this paper, \textit{convertible} denotes whether the original content contains core entities or concepts that can be clearly visualized and transformed into image-option questions while maintaining or enhancing cognitive demand. The remaining data comprised $\mathcal{D}_C$, input to the content discrimination module. Each record in $\mathcal{D}_C$ includes the context, image, and answer. From $\mathcal{D}_C$, $812$ ($40$\%) convertible instances were selected as $\mathcal{D}_Q$ for downstream tasks. Three human annotators evaluated the convertibility of $\mathcal{D}_C$, and $51.6$\% of the instances were labeled ``TRUE" in the convertible column. They also created new questions for $\mathcal{D}_Q$. A summary of the four datasets can be found in Appendix~\ref{sec:data} and Figure~\ref{fig8}.

\paragraph{Metrics} We adopt automatic and human evaluation to assess the performance of \texttt{\textbf{CmOS}}. Specifically, different tasks are evaluated using tailored metrics. For content discrimination, we use \textbf{Accuracy} to measure the its ability to identify convertible content. For question generation, we utilize \textbf{BLEU-4} \citep{10.3115/1073083.1073135} and \textbf{ROUGE-L} \citep{lin2004rouge} for question generation, both of which have been widely used in AQG works. For visual option, we adopt the Structural Similarity Index (\textbf{SSIM}) \citep{sara2019image} and the \textbf{CLIP-T} \citep{li-etal-2024-unimo} as evaluation metrics. SSIM evaluates the perceptual quality of visual options by jointly modeling luminance, contrast, and structural consistency. CLIP-T quantifies the semantic alignment between generated visual option and the corresponding description. Given the limitations of automatic evaluation in capturing human perception, we further incorporate human evaluation. The specific criteria are presented in the human evaluation section.

\paragraph{Baselines} For content discrimination and question generation, we compare our \texttt{\textbf{CmOS}} with state-of-the-art (SOTA) methods, including \textbf{VL-T5} \citep{yeh2022multi}, \textbf{MultiQG-Ti} \citep{wang2023multiqg}, \textbf{Multimodal-CoT} \citep{zhang2023multimodal}, \textbf{Chain-of-Exemplar} \citep{luo2024chain} (All these baselines, as well as our \textbf{\texttt{CmOS}}, adopt \textsc{Qwen2.5-VL-7B-Instruct} \citep{bai2023qwen} as the backbone), and \textbf{\textsc{ChatGPT}} under both zero-shot and in-context learning settings (with up to three exemplars in prompts). For visual option generation, due to the limited prior work on educational distractor visuals, we evaluate off-the-shelf model APIs (\textbf{\textsc{Flux-schnell}}, \textbf{\textsc{DALLE-3}}, \textbf{\textsc{Stable Diffusion-xl}}, \textbf{\textsc{Wanx2.1-plus}}) using identical option descriptions. Our method adopts \textsc{Qwen2.1-turbo} as the backbone. Baselines and other experimental details are provided in Appendix~\ref{sec:model}.

\subsection{Evaluation on Content Discrimination}
First, we evaluate the discrimination accuracy of \texttt{\textbf{CmOS}} and baselines in determining content convertibility. As shown in Figure~\ref{cd}, \texttt{\textbf{CmOS}}, which retrieves similar exemplars from a small pool, significantly improves the accuracy and achieves an average of 88.2\%, outperforming all baselines. Notably, although \textsc{ChatGPT} shows lower accuracy in the zero-shot, few-shot prompt significantly enhances its score. Particularly, its average accuracy under the 3 shot becomes comparable to CoE.

In terms of subject domains, all baseline models exhibit the lowest discrimination accuracy on questions from the Natural Sciences (NAT), while achieving the highest accuracy on those from the Language Sciences (LAN). Furthermore, questions accompanied by images (IMG) tend to yield lower discrimination accuracy compared to text-only questions (TXT), indicating potential challenges in multimodal reason. In addition, questions designed for students in higher grades are generally more difficult to discriminate accurately than those intended for lower grades, suggesting increased complexity in advanced educational content.

\subsection{Evaluation on Question Generation}
\paragraph{Automatic Evaluation} Table~\ref{qg} presents the performance of \texttt{\textbf{CmOS}} with $\alpha = 0.6$, compared to SOTA models in terms of BLEU-4 and ROUGE-L. The results show that \textbf{\texttt{CmOS}} significantly outperforms all baselines on both metrics, regardless of subject area, modality, or grade level. Among them, MultiQG-TI and Multimodal-CoT, which leverage MLLMs fine-tuned with CoT prompting, slightly outperform VL-T5, a pretrained language model enhanced with visual understanding. Compared to MultiQG-TI and Multimodal-CoT, CoE achieves better performance by integrating exemplar-based CoT reasoning. The table also reports \textsc{ChatGPT}'s performance under zero-shot and few-shot. Although \textsc{ChatGPT} benefits from more in-context examples, it remains substantially behind our \texttt{\textbf{CmOS}} on the multimodal question generation.

Furthermore, across the three subjects, all baselines consistently achieve the highest performance in Social Science (SOC) and the lowest in Language Science (LAN). These baselines also exhibit improved performance on image-paired questions (IMG) compared to text-only ones (TXT). In contrast, \texttt{\textbf{CmOS}} demonstrates stable performance across different subjects and grade levels, highlighting the general ability of the framework in educational content. However, BLEU-4 and ROUGE-L primarily measure surface-level lexical overlap between generated and reference questions, failing to capture semantic relevance. To address this limitation, we incorporate the METEOR metric~\citep{banerjee2005meteor}, which accounts for semantic matching, and also report the results in Appendix~\ref{METEOR}.

\begin{table*}[htbp]
\centering
\resizebox{\textwidth}{!}{%
\begin{tabular}{l|ccc|cc|cc|c}
    \toprule
    \textbf{Method} & \multicolumn{3}{c|}{\textbf{Subject}} & \multicolumn{2}{c|}{\textbf{Modality}} & \multicolumn{2}{c|}{\textbf{Grade}} & \textbf{AVG} \\
    B-4$\uparrow$ / R-L$\uparrow$ & \textbf{NAT} & \textbf{SOC} & \textbf{LAN} & \textbf{TXT} & \textbf{IMG} & \textbf{G1-6} & \textbf{G7-12} &  \\
    \midrule
    \rowcolor{white!10} 
    0 shot & $9.2 / 33.2$ & $5.0 / 17.2$ & $4.5 / 25.3$ & $3.5 / 26.2$ & $8.5 / 33.6$ & $6.7 / 30.1$ & $4.2 / 28.1$ & $4.9 / 28.3$ \\
    \rowcolor{red!10}
    1 shot & $19.5 / 37.2$ & $19.6 / 35.8$ & $10.4 / 26.2$ & $18.8 / 35.8$ & $19.6 / 38.6$ & $18.4 / 35.8$ & $21.7 / 38.6$ & $19.3 / 36.6$ \\
    3 shot & $28.6 / 46.9$ & $29.0 / 51.6$ & $27.4 / 53.6$ & $27.3 / 47.9$ & $29.8 / 48.8$ & $30.0 / 48.5$ & $27.1 / 48.6$ & $29.1 / 48.5$ \\
    \rowcolor{red!10}
    VL-T5 & $39.8 / 55.6$ & $37.3 / 45.0$ & $34.3 / 46.1$ & $36.1 / 50.2$ & $40.7 / 52.7$ & $38.1 / 50.9$ & $39.6 / 54.9$ & $39.1 / 51.5$ \\
    MultiQG-TI & $43.2 / 58.7$ & $39.8 / 47.2$ & $37.3 / 49.0$ & $38.9 / 52.7$ & $43.5 / 55.9$ & $41.0 / 53.7$ & $42.0 / 58.1$ & $42.3 / 55.0$ \\
    \rowcolor{red!10}
    Multimodal-CoT & $47.6 / 63.3$ & $44.4 / 53.2$ & $42.9 / 53.8$ & $44.0 / 57.4$ & $48.3 / 60.9$ & $47.0 / 58.9$ & $47.7 / 63.0$ & $46.6 / 59.8$ \\
    CoE & $55.7 / 72.3$ & $52.7 / 61.5$ & $46.9 / 57.3$ & $51.6 / 66.7$ & $56.0 / 69.9$ & $54.2 / 67.7$ & $54.8 / 72.1$ & $54.7 / 69.9$ \\
    \rowcolor{red!10}
    \textbf{\texttt{CmOS}} & $\underline{\textbf{76.8}} / \underline{\textbf{77.5}}$ & $\underline{\textbf{81.1}} / \underline{\textbf{79.2}}$ & $\underline{\textbf{50.5}} / \underline{\textbf{63.2}}$ & $\underline{\textbf{75.6}} / \underline{\textbf{76.7}}$ & $\underline{\textbf{78.6}} / \underline{\textbf{78.4}}$ & $\underline{\textbf{79.3}} / \underline{\textbf{78.4}}$ & $\underline{\textbf{70.1}} / \underline{\textbf{74.5}}$ & $\underline{\textbf{75.5}} / \underline{\textbf{77.2}}$ \\
    \hline
    \hspace{1em}w /o Discriminator & $71.6 / 71.7$ & $77.2 / 74.2$ & $49.5 / 60.6$ & $72.3 / 71.3$ & $73.9 / 72.5$ & $74.3 / 73.9$ & $69.8 / 71.7$ & $72.9 / 72.1$ \\
    \rowcolor{red!10}
    \hspace{1em}w /o OQRM & $44.8 / 66.5$ & $41.3 / 55.9$ & $37.6 / 57.2$ & $39.1 / 61.1$ & $43.7 / 60.6$ & $42.1 / 59.8$ & $42.6 / 65.9$ & $42.4 / 62.3$ \\
    \bottomrule
\end{tabular}
}
\caption{Automatic evaluation results of question generation. $\uparrow$: higher is better.}
\label{qg}
\end{table*}

\paragraph{Human Evaluation} In addition to automatic evaluation, we conduct human evaluation to further assess the quality of generated questions. We randomly sample $50$ questions from different methods and recruit three annotators to rate each question on a 1-5 scale across four criteria: (1) \textbf{Fluency} \citep{song2016question}, assessing the naturalness and readability of the question; (2) \textbf{Grammaticality} \citep{heilman2011automatic}, measuring syntactic correctness; (3) \textbf{Complexity} \citep{rodriguez2022end}, evaluating the cognitive or linguistic challenge posed by the question; and (4) \textbf{Relevance} \citep{chughtai2022lecture}, measuring how well the question matches the background content. Detailed guidelines are provided in Appendix~\ref{gl}. 

As shown in Table~\ref{tab:question_generation_human}, although the ground-truth questions achieve the highest scores across all metrics, \texttt{\textbf{CmOS}} outperforms all baselines and obtains the closest performance to the ground-truth, with an average score of $4.58$. These results demonstrate that our framework can generate high-quality educational questions that are fluent,  correct, engaging, and relevant to the content. Moreover, both Multimodal-CoT and CoE significantly outperform VL-T5, highlighting the effectiveness of MCoT reasoning. Although \textsc{ChatGPT} trails \texttt{\textbf{CmOS}} in complexity and relevance, it surpasses VL-T5 in fluency and grammaticality, indicating its strong ability to generate well-formed natural language context.
\begin{table}[htbp]
\centering
\resizebox{\columnwidth}{!}{%
\begin{tabular}{l|cccc|c}
    \toprule
    \textbf{Method} & \textbf{Flu.}$\uparrow$ & \textbf{Gra.}$\uparrow$ & \textbf{Com.}$\uparrow$ & \textbf{Rel.}$\uparrow$ & \textbf{AVG}$\uparrow$ \\
    \midrule
    \rowcolor{white!10} 
    ChatGPT4 & $4.65$ & $4.71$ & $3.18$ & $3.24$ & $3.94$ \\
    \rowcolor{red!10} 
    VL-T5 & $4.31$ & $4.56$ & $3.49$ & $3.55$ & $3.97$ \\
    \rowcolor{white!10} 
    MultiQG-TI & $4.61$ & $4.65$ & $3.99$ & $4.12$ & $4.34$ \\
    \rowcolor{red!10} 
    Multimodal-CoT & $4.65$ & $4.68$ & $4.04$ & $4.25$ & $4.41$ \\
    \rowcolor{white!10} 
    CoE & $4.65$ & $4.72$ & $4.25$ & $4.29$ & $4.48$ \\
    \rowcolor{red!10} 
    \textbf{\texttt{CmOS}} & \underline{\textbf{4.69}} & \underline{\textbf{4.78}} & \underline{\textbf{4.37}} & \underline{\textbf{4.49}} & \underline{\textbf{4.58}} \\
    \hline
    \rowcolor{white!10} 
    Groundtruth & $4.72$ & $4.82$ & $4.59$ & $4.57$ & $4.68$ \\
    \bottomrule
\end{tabular}
}
\caption{Human evaluation results of question generation. $\uparrow$: higher is better.}
\label{tab:question_generation_human}
\end{table}

\subsection{Evaluation on Visual Option Generation}
\paragraph{Automatic Evaluation} Table~\ref{og} presents automatic evaluation results for visual option generation with $\beta=1.4$. \texttt{\textbf{CmOS}} significantly outperforms four baselines in SSIM and CLIP-T by integrating MLLM with T2I model. These results suggest that \texttt{\textbf{CmOS}} effectively improves both the structural similarity among visual options and their semantic alignment with descriptions. Benefiting from strong semantic understanding and stylistic generalization, \textsc{Wanx2.1-plus} slightly outperforms the other three baselines in most categories. In contrast, no substantial differences are observed among the remaining baselines. The hyperparameter analysis results for $\alpha$ and $\beta$ are provided in Appendix~\ref{hp}.

To further evaluate performance across disciplines, we analyze the results for three academic subjects. Remarkably, \texttt{\textbf{CmOS}} achieves the best performance in the social sciences (SOC), but performs worst in language sciences (LAN), mirroring the trends observed in question generation performance. Moreover, for image-equipped (IMG) questions, \texttt{\textbf{CmOS}} obtains the highest scores in both SSIM and CLIP-T, while for text-only (TXT) questions, its performance degrades significantly on both metrics. The findings demonstrate that visual input enhances the semantic fidelity and contextual plausibility of generated descriptions. Additionally, the consistently strong performance of our \texttt{\textbf{CmOS}} framework across subjects and grade levels further demonstrates its robust generalization ability.

\begin{table*}[htbp]
\centering
\resizebox{\textwidth}{!}{%
\begin{tabular}{l|ccc|cc|cc|c}
    \toprule
    \textbf{Method} & \multicolumn{3}{c|}{\textbf{Subject}} & \multicolumn{2}{c|}{\textbf{Modality}} & \multicolumn{2}{c|}{\textbf{Grade}} & \textbf{AVG} \\
    SSIM$\uparrow$ / CLIP-T$\uparrow$ & \textbf{NAT} & \textbf{SOC} & \textbf{LAN} & \textbf{TXT} & \textbf{IMG} & \textbf{G1-6} & \textbf{G7-12} &  \\
    \midrule
    \rowcolor{white!10} 
    Flux.schnell & $39.0 / 28.9$ & $41.5 / 29.1$ & $29.7 / 29.4$ & $37.6 / 29.4$ & $42.2 / 28.2$ & $39.2 / 29.2$ & $38.8 / 28.5$ & $39.1 / 29.0$ \\
    \rowcolor{blue!10} 
    DALLE-3 & $40.2 / 30.1$ & $43.8 / 29.3$ & $30.6 / 29.0$ & $38.4 / 30.2$ & $43.1 / 29.0$ & $42.4 / 29.9$ & $39.5 / 29.3$ & $40.2 / 29.7$ \\
    StableDiffusion-XL & $40.6 / 27.3$ & $43.9 / 28.5$ & $31.4 / 27.4$ & $39.4 / 28.4$ & $43.5 / 27.1$ & $41.2 / 28.1$ & $40.8 / 27.5$ & $40.7 / 27.9$ \\
    \rowcolor{blue!10} 
    Wanx2.1-plus & $41.8 / 31.6$ & $44.9 / 30.2$ & $32.1 / 28.5$ & $40.6 / 30.4$ & $44.7 / 32.3$ & $42.1 / 31.2$ & $40.8 / 30.1$ & $41.5 / 30.8$ \\
    \textbf{\texttt{CmOS}}(Wanx2.1-turbo) & $\textbf{\underline{59.0}} / 40.6$ & $\textbf{\underline{61.2}} / 42.8$ & $\textbf{\underline{49.7}} / \textbf{\underline{37.6}}$ & $\textbf{\underline{58.3}} / 38.4$ & $\textbf{\underline{62.1}} / 42.7$ & $\textbf{\underline{59.7}} / 40.7$ & $\textbf{\underline{59.1}} / 39.5$ & $\textbf{\underline{59.5}} / 40.2$ \\
    \hline
    \rowcolor{blue!10} 
    \hspace{1em}w / o Discriminator & $58.2 / 29.3$ & $59.4 / 30.9$ & $49.1 / 25.6$ & $54.8 / 26.9$ & $60.3 / 31.1$ & $57.9 / 29.3$ & $56.8 / 27.7$ & $57.5 / 28.8$ \\
    \hspace{1em}w / o Reasoning & $50.8 / 38.0$ & $58.8 / 40.3$ & $42.0 / 36.3$ & $50.9 / 37.6$ & $54.0 / 41.3$ & $52.5 / 39.6$ & $51.7 / 38.0$ & $52.3 / 39.1$ \\
    \rowcolor{blue!10} 
    \hspace{1em}w / o Template & $48.0 / \textbf{\underline{41.5}}$ & $50.7 / \textbf{\underline{44.6}}$ & $38.8 / 36.9$ & $47.4 / \textbf{\underline{40.1}}$ & $50.8 / \textbf{\underline{43.3}}$ & $48.9 / \textbf{\underline{41.6}}$ & $47.2 / \textbf{\underline{40.3}}$ & $48.3 / \textbf{\underline{41.1}}$\\
    \hspace{1em}w / o Tuning & $54.8 / 31.7$ & $57.7 / 35.2$ & $44.8 / 29.5$ & $53.9 / 30.5$ & $58.1 / 35.4$ & $55.4 / 33.5$ & $54.7 / 31.4$ & $55.1 / 32.7$ \\
    \bottomrule
\end{tabular}
}
\caption{Automatic evaluation results of visual option generation. $\uparrow$: higher is better.}
\label{og}
\end{table*}

\paragraph{Human Evaluation} Similarly, we invited three qualified annotators to subjectively evaluate the visual options generated by different methods. Using a 5-point Likert scale, they rated each visual option across three important dimensions: (1) \textbf{Plausibility} \citep{luo2024chain}, assessing coherence with the background content and question context; (2) \textbf{Distractibility} \citep{gierl2017developing}, measuring the cognitive burden posed by distractor visual options; and (3) \textbf{Engagement} \citep{gierl2017developing}, reflecting how much the visual options attract learners' attention. Guidelines are detailed in Appendix~\ref{gl}.

Table~\ref{tab:human_evaluation_distractor} presents a summary of the human evaluation results. Generally, \textbf{\texttt{CmOS}} surpasses other T2I models in terms of plausibility and distractibility. This indicates that the visual options it generates are more semantically consistent and possess a higher level of misleadingness, sufficient to test human judgment. However, even though image templates and tuning was carried out to enhance engagement, the improvement is relatively limited. \textbf{\texttt{CmOS}} performs slightly worse than \textsc{DALLE}-3 but marginally better than \textsc{StableDiffusion-xl}. It is worth noting that the average scores across all methods are relatively low. Specifically, \textbf{\texttt{CmOS}} only achieves a score of $3.55$ out of $5$. This situation highlights that creating visual content with pedagogical effectiveness still poses a large challenge.

\begin{table}[htbp]
\centering
\resizebox{\columnwidth}{!}{
    \begin{tabular}{l|ccc|c}
        \toprule
        \textbf{Method} & \textbf{Plaus.}$\uparrow$ & \textbf{Distra.}$\uparrow$ & \textbf{Enga.}$\uparrow$ & \textbf{AVG}$\uparrow$ \\
        \midrule
        Flux-schnell & $3.07$ & $2.99$ & $3.90$ & $3.32$ \\
        \rowcolor{blue!10} 
        DELLE-3 & $3.13$ & $2.86$ & $\textbf{\underline{4.20}}$ & $3.41$ \\
        StableDiffusion-XL & $3.24$ & $2.82$ & $4.13$ & $3.40$ \\
        \rowcolor{blue!10} 
        Wanx2.1-plus & $3.14$ & $2.75$ & $4.11$ & $3.33$ \\
        \hline
        \textbf{\texttt{CmOS}}(Wanx2.1-turbo) & $\textbf{\underline{3.51}}$ & $\textbf{\underline{3.09}}$ & $4.17$ & $\textbf{\underline{3.55}}$ \\
        \bottomrule
    \end{tabular}
}
\caption{Human evaluation results of visual option generation. $\uparrow$: higher is better.}
\label{tab:human_evaluation_distractor}
\end{table}

\subsection{Ablation Study}
We perform ablation studies to investigate the effects of the proposed approaches in terms of similar exemplar retrieval, optimal question-reason match, template and tuning, as presented in Figure \ref{cd}, Table \ref{qg} and Table \ref{og}. There are several notable findings.  

\textbf{Finding1:} Figure~\ref{cd} illustrates that removing the retrieval of similar exemplars forces the model to rely solely on abstract judgment criteria when assessing whether input content can be converted into visual-option questions. This change causes accuracy to drop by $19.5$\%, suggesting that concrete reason exemplars contribute more effectively to content discrimination than predefined standards.

\textbf{Finding2:} In question generation, as shown in Table \ref{qg}, removing the discriminator leads to BLEU-4 and ROUGE-L drops of $2.6$ and $5.1$, reflecting weaker semantic alignment and greater inconsistency. This indicates that unconvertible inputs disrupt downstream processing, thereby diminishing coherence and overall quality. For visual option generation, as shown in Table \ref{og}, SSIM remains largely unchanged, whereas CLIP-T decreases by $11.4$, the largest drop among ablations. These results indicate that unsuitable inputs hinder the T2I model’s ability to align images with text.

\textbf{Finding3:}  Table~\ref{qg} clearly reveals that excluding the OQRM module significantly weakens question generation, with BLEU-4 plummeting by $33.1$ and ROUGE-L dropping by $7.3$. This sharp decline underscores OQRM’s importance in identifying question-reason pairs with strong lexical and semantic fidelity to human-written references.

\textbf{Finding4:} As shown in Table \ref{og}, when the reason generator is removed and distractors are produced solely from the question and answer, both SSIM and CLIP-T scores decline to varying extents. Specifically, SSIM decreases by $7.2$, while CLIP-T shows a slight drop of $1.1$. This indicates that the inclusion of reasons primarily enhances the visual similarity among options, with comparatively smaller effects on image-text alignment.

\textbf{Finding5:} As shown in Table~\ref{og}, removing the template results in an $11.2$ drop in SSIM, while CLIP-T scores unexpectedly improve. This divergence suggests that templates enhance visual consistency across options but may impose constraints that hinder semantic alignment with textual descriptions. Further ablation reveals that disabling tuning by MLLM and T2I model consistently leads to notable declines in both SSIM ($-4.4$) and CLIP-T ($-12.5$), reinforcing the importance of iterative tuning strategies for improving intra-option visual similarity and cross-modal coherence.

\subsection{Case Study}

\begin{figure}[t]
  \includegraphics[width=\columnwidth] {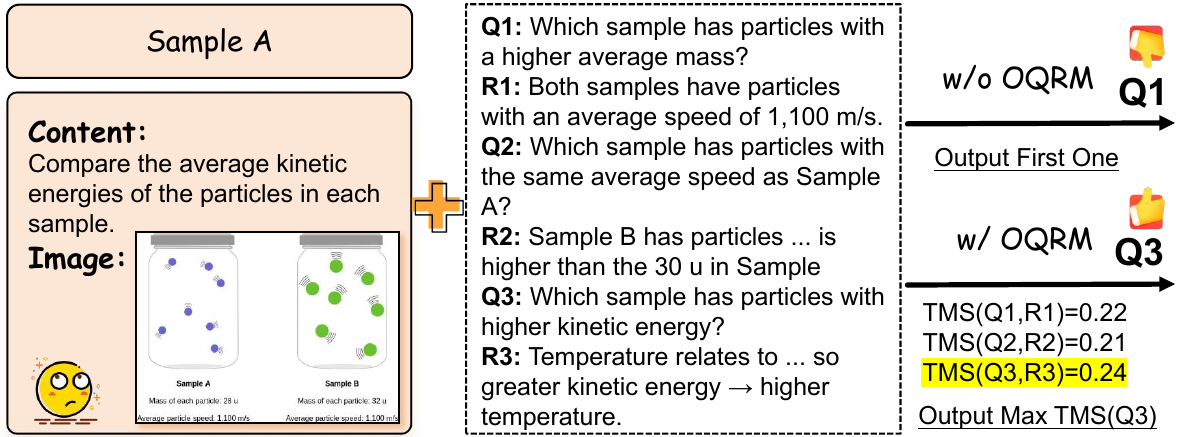}
  \caption{Case in terms of the OQRM module.}
  \label{case1}
\end{figure}

\begin{figure}[t]
  \includegraphics[width=\columnwidth] {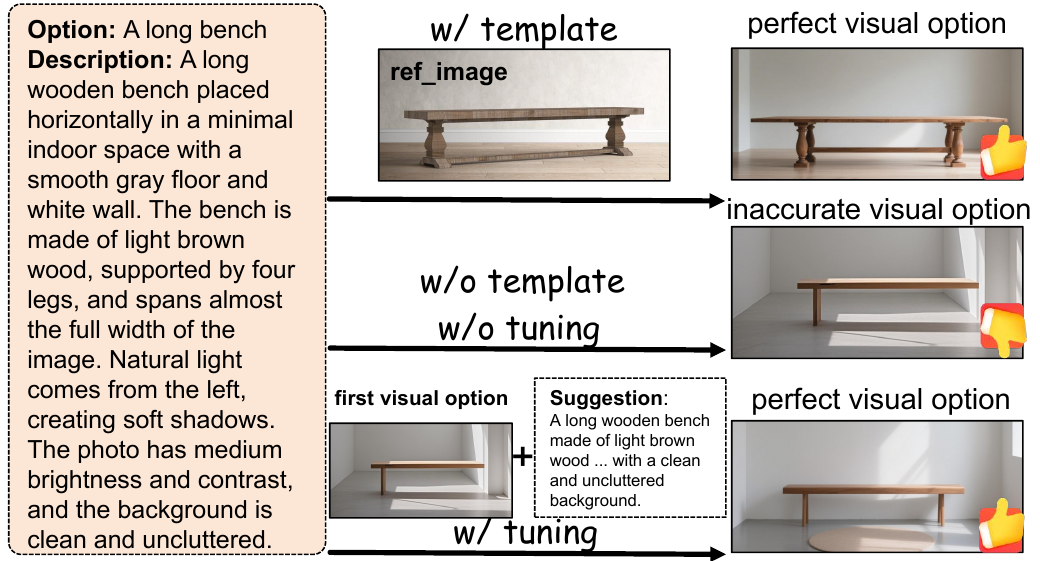}
  \caption{Case regarding the Template and Tuning.}
  \label{case2}
\end{figure}

To qualitatively assess the three important modules in \textbf{\texttt{CmOS}}, Figures \ref{case1} and \ref{case2} present representative examples from ScienceQA, which include the context, an image, and an answer. Without computing Total Match Scores (TMS) across multiple candidate question-reason pairs to select the one with the highest alignment, the generator tends to generate the most probable question, which often leads to suboptimal performance in question generation.

To further examine the effect of the template and tuning modules on visual option generation, a comparative example are shown in Figure \ref{case2}. When the template module is incorporated, the generated visual options exhibit improved semantic plausibility and consistency with object properties. In contrast, removing both the template and tuning modules results in outputs that deviate from commonsense expectations and display stylistic inconsistency. Additionally, the example illustrates that the tuning process enables iterative refinement: even if the initial visual option is inadequate, tuning can adjust and improve it toward the desired quality. These findings suggest that the template module ensures baseline plausibility, while the tuning supports optimization, and their combination contributes to overall improvements in visual option quality.
\section{Conclusion}
In this paper, we present a novel framework called
\textbf{C}ross-\textbf{m}odal \textbf{O}ptions \textbf{S}ynthesis (\texttt{\textbf{CmOS}}), which combines retrieved similar exemplars and Multimodal Chain-of-Thought (MCoT) reasoning to generate educational multiple-choice questions and visual options from multimodal input. Specifically, we utilize MLLMs to encode multimodal contexts and incorporate them into a three-stage Multimodal-CoT framework, namely content discrimination, question generation, and visual option generation. Meanwhile, we introduce a similar exemplar retrieval module to guide the discrimination. What's more, we use OQRM module to select optimal question and reasoning progress. Finally, we leverage template-based and 
slight tuning strategies to generate educational visual options. Our experimental results on three test sets demonstrate that \texttt{\textbf{CmOS}} outperforms all existing methods and models, achieving new state-of-the-art performance. More importantly, our study highlights the potential of visual-option-based multiple-choice question generation to enhance multimodal teaching resources, support personalized learning, and foster deeper understanding in educational settings.
\section*{Limitations}
Despite its promising performance, our proposed framework still has two limitations.

\paragraph{Exemplar Resource} On one hand, similar to CoE framework, our similar exemplar-based strategy for enhancing content discrimination accuracy has an inherent limitation: the dependency on a fixed pool of exemplars. When the input content falls outside the distribution of datasets like ScienceQA, the retrieved exemplars may exhibit low semantic similarity, leading to degraded discrimination performance. To mitigate this, future work could explore adaptive retrieval mechanisms or augment the exemplar pool with more diverse, domain-general instances, possibly using synthetic data or continual learning techniques to improve generalization beyond the source domain.

\paragraph{Visual Option Quality} On the other hand, despite improvements in text-image alignment through image templates and lightweight tuning, the generated visual options still suffer from limited visual detail, occasional content hallucinations, and inconsistent style. Human evaluation (Table~\ref{tab:human_evaluation_distractor}) shows relatively low scores in plausibility and distractibility, indicating that some generated visual options are semantically weak, visually indistinct, or pedagogically uninformative, thereby limiting their effectiveness in supporting deep cognitive processing or engaging prior knowledge. These issues stem from the use of general-purpose image generation models that are not optimized for educational applications, often resulting in irrelevant, ambiguous, or stylistically incoherent outputs. Future work could fine-tune generation models on curated educational datasets and apply task-specific constraints or multimodal alignment objectives to improve clarity, visual contrast, semantic accuracy, and pedagogical value.

\section*{Ethics Statement}
We comply with institutional ethical guidelines throughout this study. No private or non-public data was used. For human annotation (Sections 4.3 and 4.4), six annotators were recruited from the schools of education at local universities via public advertisements, with clear disclosure of compensation terms. All annotators were senior undergraduate or graduate students in education-related programs who participated on a part-time basis. Each annotator was compensated at a rate of $55$ CNY per hour, which exceeds the local minimum hourly wage for part-time employment in 2025 ($23.5$ CNY/hour). The annotation process did not involve any personally sensitive information.

\bibliography{main}

\appendix
\section{Related Works}
\label{rw}
Educational question is an indispensable component of instructional resources, serving multiple functions including assessment, guidance, feedback, and the promotion of active learning \citep{tofade2013best,thalheimer2014learning}. However, manually authoring educational questions is a complex and resource-intensive task that requires professional training, domain knowledge, and instructional experience \citep{davis2009tools,kim2012incorporation}. To address the high cost and inefficiency associated with manual question generation, Automatic Question Generation (AQG) technologies have emerged as a promising solution \citep{brown2005automatic}, and have been widely applied in dialogue systems \citep{gao2019interconnected,gu2021chaincqg,bulathwela2023scalable} and intelligent tutoring systems \citep{kulshreshtha2022few,xu2022fantastic,yadav2023contextualizing}, becoming a prominent research focus within the field of Artificial Intelligence in Education to support personalized learning \citep{bulathwela2023scalable,fawzi2024towards,lamsiyah2024fine}.

\section{Dataset Details}

   
Figure \ref{fig8} shows four datasets distribution across 3 subjects (NAT, SOC, LAN), 2 modalities (IMG, TXT), and 2 grade ranges (G1-6, G7-12). Question types: NAT = natural science, SOC = social science, LAN = language science, TXT = only containing text, IMG = containing image, G1-6 = grades 1-6, G7-12 = grades 7-12. For $\mathcal{D}_O$, NAT has $2,252$, SOC $1,100$, and LAN $889$; IMG has $2,224$ and TXT $2,017$; G1-6 has $2,723$ and G7-12 has $1,518$. For $\mathcal{D}_E$, NAT has $257$, SOC $128$, and LAN $97$; IMG has $228$ and TXT $254$; G1-6 has $309$ and G7-12 has $173$. For $\mathcal{D}_C$, NAT has $1,990$, SOC $969$, and LAN $787$; IMG has $1,795$ and TXT $1,964$; G1-6 has $2,723$ and G7-12 has $1,036$. For $\mathcal{D}_Q$, NAT has $576$, SOC $236$; IMG has $571$ and TXT $241$; G1-6 has $568$ and G7-12 has $244$.

\label{sec:data}
\section{Implementation Details}
\label{sec:model}

To ensure fair and reproducible evaluation of visual option generation baselines, we report the detailed settings of all off-the-shelf generative models used in our experiments. As part of our pipeline, we employ Qwen2.5-VL-7B-Instruct \citep{bai2023qwen} for content discrimination, question generation, and the production of visual option descriptions. Moreover, we uniformly set the decoding parameters to \texttt{top-p} = $0.8$ and \texttt{temperature} = $0.7$. To generate corresponding images, we use Wanx2.1-turbo as our main image synthesis backbone. Additionally, we utilize the robust Contrastive Language–Image Pretraining model FARE \citep{schlarmann2024robust} for retrieval and evaluation. Below, we present the configurations of these models and the prompting details for MCoT. 

\subsection*{Models and Platforms}

\subsubsection*{(1) \textsc{Flux-schnell}} 
\begin{itemize}
  \item \texttt{Access}: Alibaba Bailian platform\footnote{\url{https://bailian.console.aliyun.com}} 
  \item \texttt{guidance\_scale}: $3.5$
  \item \texttt{num\_inference\_steps}: $50$
  \item \texttt{image\_size}: $1024\times1024$
  \item \texttt{seed}: $42$
\end{itemize}

\subsubsection*{(2) \textsc{DALLE-3}}
\begin{itemize}
  \item \texttt{Access}: OpenAI official API\footnote{\url{https://platform.openai.com}}
  \item \texttt{model}: dall-e-3
  \item \texttt{num\_inference\_steps}: N/A
  \item \texttt{image\_size}: $1024\times1024$
  \item \texttt{seed}: N/A
\end{itemize}

\subsubsection*{(3) \textsc{StableDiffusion-XL}} 
\begin{itemize}
  \item \texttt{Access}: Alibaba Bailian platform
  \item \texttt{guidance\_scale}: $10$
  \item \texttt{num\_inference\_steps}: $50$
  \item \texttt{image\_size}: $1024\times1024$
  \item \texttt{seed}: N/A
\end{itemize}

\subsubsection*{(4) \textsc{Wanx2.1-plus}} 
\begin{itemize}
  \item \texttt{Access}: Alibaba Bailian platform
  \item \texttt{guidance\_scale}: N/A
  \item \texttt{num\_inference\_steps}: N/A
  \item \texttt{image\_size}: $1024\times1024$
  \item \texttt{seed}: $42$
  \item \texttt{negative\_prompt} = N/A
\end{itemize}

\subsubsection*{(5) \textsc{Wanx2.1-turbo}} 
\begin{itemize}
  \item \texttt{Access}: Alibaba Bailian platform
  \item \texttt{guidance\_scale}: N/A
  \item \texttt{num\_inference\_steps}: N/A
  \item \texttt{image\_size}: $1024\times1024$
  \item \texttt{seed}: $42$
  \item \texttt{negative\_prompt} = N/A
\end{itemize}

\subsection*{Prompts}
The first red-shaded panel presents the prompt used to guide \textsc{Qwen2.5-vl-72B} for exemplar construction. The second blue-highlighted section shows the prompts employed to instruct \texttt{\textbf{CmOS}}, VL-T5, MultiQG-TI, MultiModal-CoT, and CoE in content discrimination, question and reason generation, textual option generation, visual description generation, and visual option optimization. Note that the last three tasks are exclusive to \texttt{\textbf{CmOS}}. The third yellow-marked panel provides the prompts used to instruct \textsc{ChatGPT} for content discrimination and question generation under both zero-shot and few-shot settings (CD = Content Discrimination, QG = Question Generation).

\begin{tcolorbox}[enhanced jigsaw,breakable,pad at break*=1mm,
colback=red!5!white,colframe=red!75!black,
title=Box 1: Prompt input for exemplar construction,
watermark color=white]

Context: ...\\
Image: ...\\
Answer: ...\\
Please analyze whether the above content can be transformed into a multiple-choice question with images as options, based on the following three dimensions: \\
(1) Whether the answer itself is suitable for visual transformation; \\
(2) Whether the key entities in the context are suitable for visual transformation; \\
(3) Which form of transformation (if any) provides greater educational value, or whether neither form is suitable or meaningful in an educational context.\\

Reasoning: ...\\
Convertible: ...

\end{tcolorbox}

\begin{tcolorbox}[enhanced jigsaw,breakable,pad at break*=1mm,colback=blue!5!white,colframe=blue!75!black,title=Box 2: Prompt input for \textbf{\texttt{CmOS}} and baselines,watermark color=white]
\textbf{Content Discrimination Prompt Input}
Context: ... \\
Image: ... \\
Answer: ... \\
Refer to the following exemplar to determine whether the original content can be converted into a question format with visual options and give the reason.\\
Exemplar\\
Context: ... \\
Image: ... \\
Answer: ... \\
Reason: ... \\
Judgment: ...\\

\textbf{Question Generation Prompt Input} \\
Context: ... \\
Image: ... \\
Answer: ... \\
Judgment: ... \\
Refer to the following three exemplars. Generate a new question suitable for visual options basing on the original content, provide the corresponding answer and reason. \\
Exemplar1 \\
Context: ... \\
Image: ... \\
Answer: ... \\
Question: ... \\
Reason: ...\\
Exemplar2 \\
Exemplar3 \\
\textbf{Option Generation Prompt Input} \\
Context: ... \\
Question: ... \\
Answer: ... \\
Reason: ... \\
Refer to the following exemplar content to generate multiple options and description that are related to the answer and have a certain degree of interference. \\
Exemplar \\
Context: ... \\
Question: ... \\
Answer: ... \\
Reason: ... \\
Options: (a) ...; (b) ...; (c) ...; (d) ...\\

\textbf{Visual Option Generation Prompt Input} \\
Option: a picture of ×××. \\
Description: Color; Green; Shape. \\
Please refer to the following reference image, Generate a corresponding image according to the visual description of this option.\\

\textbf{Optimization Prompt Input}\\
Visual Option: ... \\
Description: Color; Style; Shape. \\
Please calculate the similarity between the given visual option and the descriptive text, and provide optimization suggestions.\\
Reference Image: ...

\end{tcolorbox}

\begin{tcolorbox}[enhanced jigsaw,breakable,pad at break*=1mm, colback=yellow!5!white, breakable, colframe=yellow!50!black, colbacktitle=yellow!75!black, title=Box 3: Zero-shot and few-shot settings for \textsc{ChatGPT}]
  \textbf{0 shot CD and QG}\\
  Context: ... \\
  Answer: ... \\
  Image: ... \\
  Determine whether this content can be converted into a visual option question. If it is convertible, generate a question based on the corresponding content.\\
  
  \textbf{1 shot CD and QG}\\
  Context: ... \\
  Answer: ... \\
  Image: ... \\
  Refer to the exemplar, determine whether this content can be converted into a visual option question. If it is convertible, generate a question based on the corresponding content. \\
  Exemplar: ... \\
  
  \textbf{3 shot CD and QG}\\
  Context: ... \\
  Answer: ... \\
  Image: ... \\
  Refer to these 3 exemplars, determine whether this content can be converted into a visual option question. If it is convertible, generate a question based on the corresponding content. \\
  Exemplar 1: ... \\
  Exemplar 2: ... \\
  Exemplar 3: ...
\end{tcolorbox}

\section{METEOR}
Unlike BLEU-4 and ROUGE-L, which focus on lexical overlap, METEOR (MTR) also captures semantic and content-level similarities. Table~\ref{mt} shows that \texttt{\textbf{CmOS}} consistently outperforms SOTA methods in question generation under the METEOR metric. While its score in the language sciences (LAN) is lower than in natural (NAT) and social sciences (SOC), it still significantly exceeds other methods. This suggests that \texttt{\textbf{CmOS}} generates semantically aligned questions, aided by its multi-question filtering strategy.

Moreover, MultiQG-TI and Multimodal-CoT that are fine-tuned with CoT prompting, achieve better performance than VL-T5. Although VL-T5 benefits from stronger visual understanding, it lags behind in semantic coherence during question generation. In contrast, CoE leverages Chain of Exemplar reasoning to further enhance its generation quality, outperforming both MultiQG-TI and Multimodal-CoT. Additionally, the table includes results for \textsc{ChatGPT} under zero-shot and few-shot settings. While \textsc{ChatGPT}  benefits from increased contextual exemplars and achieves METEOR scores that approach those of some baselines, it still falls considerably short of \texttt{\textbf{CmOS}} in multimodal question generation, highlighting its limitations in complex reasoning and visual-semantic integration.

Similarly, we evaluated the performance of \texttt{\textbf{CmOS}}  after removing the Optimal Question-Reason Match (OQRM) module. A noticeable performance drop was observed, with the average METEOR score decreasing by 17.3 points. This result highlights the importance of OQRM in enhancing the semantic alignment between generated questions and questions authored by humans.

\label{METEOR}
\section{Hyperparameter Analysis}
\label{hp}
\subsection*{Question Generation}
\begin{figure}[t]
  \includegraphics[width=\columnwidth] {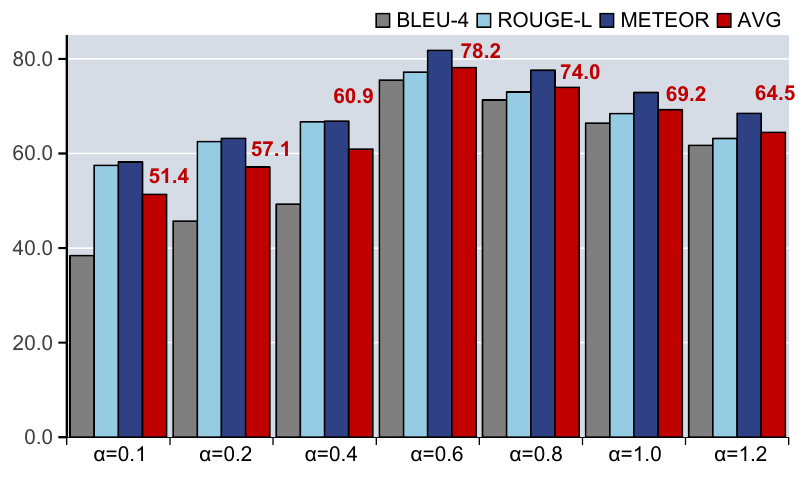}
  \caption{The overall performance of question generation with varying $\alpha$.}
  \label{fig6}
\end{figure}
To determine the optimal hyperparameter $\alpha$ for selecting the best question-reason pair, we systematically examined the BLEU and ROUGE-L scores across varying values of $\alpha$ from $0.1$ to $1.2$ in steps of $0.1$ or $0.2$. What's more, we sampled 300 instances from \( D_c \) where the value of "convertible" is true, ensuring no overlap with the dataset \( D_Q \). As shown in Figure~\ref{fig6}, both scores exhibit a trend of first increasing and then decreasing. Notably, BLEU-4, ROUGE-L and METEOR reach their peak values when $\alpha = 0.6$. The average of the two scores also reaches its highest value of 78.2 at this point. Therefore, we select $\alpha = 0.6$ as the optimal value.

\subsection*{Visual Option Generation}
\begin{figure}[t]
  \includegraphics[width=\columnwidth] {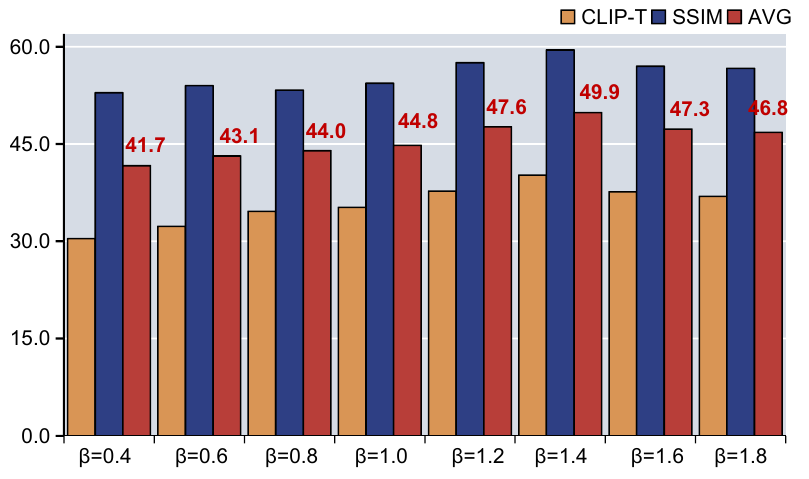}
  \caption{The overall performance of question generation with varying $\beta$.}
  \label{fig7}
\end{figure}
To determine the optimal hyperparameter $\beta$ for balancing the influence of the image itself and its caption during template retrieval, we systematically evaluated the changes in structural similarity (SSIM) and text-image similarity (CLIP-T) scores for the generated visual options across different values of $\beta$. As shown in the Figure \ref{fig7}, $\beta$ was gradually increased from $0.4$ to $1.8$ in increments of $0.2$. The results indicate that both SSIM and CLIP-T scores exhibit a trend of initially increasing and then decreasing as $\beta$ increases. Specifically, when $\beta = 1.4$, the SSIM score reaches its peak at $59.3$, and the CLIP-T score also achieves its highest value of $40.4$. Therefore, we select $\beta = 1.4$ as the optimal value within the range of our experimental settings for the subsequent visual option generation task.
\section{Analysis of Question Diversity}
We adopt Distinct-n scores to evaluate the diversity of questions generated. Specifically, this metric calculates the number of unique n-grams at the corpus level, where higher values indicate greater diversity. We consider values of n ranging from $1$ to $4$. As shown in the table \ref{tab:distinct}, overall performance improves with increasing n. Both \textsc{ChatGPT}'s 0-shot and few-shot settings exhibit relatively high question diversity. Aside from \textsc{ChatGPT}, \textbf{\texttt{CmOS}} only falls slightly behind CoE on Distinct-1, while it surpasses the baseline methods on other three metrics. When compared to the Groundtruth, we find that both CoE and \texttt{\textbf{CmOS}} achieve scores close to human-generated questions, suggesting that these two methods better approximate the style and distribution of human-authored question diversity.
\begin{table}[ht]
\centering
\resizebox{\columnwidth}{!}{
\begin{tabular}{lcccc}
\toprule
\textbf{Method} & \textbf{Distinct-1} & \textbf{Distinct-2} & \textbf{Distinct-3} & \textbf{Distinct-4} \\
\midrule
\rowcolor{blue!10}
0shot          & $19.68$ & $38.32$ & $49.89$ & $58.81$ \\
1shot          & $18.46$ & $34.16$ & $44.49$ & $53.32$ \\
\rowcolor{blue!10}
3shot          & $17.21$ & $30.80$ & $39.95$ & $47.93$ \\
VL-T5          & $12.11$ & $23.95$ & $30.93$ & $37.74$ \\
\rowcolor{blue!10}
MultiQG-TI     & $13.39$ & $24.54$ & $31.63$ & $38.02$ \\

Multimodal-CoT & $15.92$ & $23.31$ & $36.20$ & $40.47$ \\
\rowcolor{blue!10}
CoE            & $17.20$ & $29.47$ & $38.18$ & $45.74$ \\
\textbf{\texttt{CmOS}} & $16.85$ & $34.65$ & $40.72$ & $47.22$ \\
\rowcolor{blue!10}
Grondtruth     & $17.41$ & $31.56$ & $40.46$ & $47.97$ \\
\bottomrule
\end{tabular}}
\caption{Distinct-n results of different methods.}
\label{tab:distinct}
\end{table}

\section{Analysis of Different Base Models}
To analyse the generality of \texttt{\textbf{CmOS}}, we conduct an experiment to utilize other base models in place of \textsc{Qwen2.5-VL-7B-Instruct} as the backbone for question and visual option generation, including \textsc{Llama3.2-11B-Vision} \citep{grattafiori2024llama3herdmodels}, \textsc{LLaVA} \citep{lu2022learn}, \textsc{InstructBLIP} \citep{dai2023instructblipgeneralpurposevisionlanguagemodels}, \textsc{mPLUG-Owl} \citep{ye2024mplugowlmodularizationempowerslarge}, and \textsc{VisualGLM-6B} \citep{du2022glmgenerallanguagemodel}. Note that we employ the same prompt for all base models to ensure fairness in the comparison. As summarized in Table 9, \textsc{Qwen2.5-VL-7B-Instruct} outperforms all the rest base models, showcasing its high applicability and suitability in our framework. Generally, while there are slight difference in performance among the 6 base models, they consistently demonstrate superior performance in both question and visual option generation, which further confirms the effectiveness and versatility of our \texttt{\textbf{CmOS}} framework.

\begin{table}[htbp]
\small
\centering
\resizebox{\columnwidth}{!}{
    \begin{tabular}{l|ccc|cc}
        \toprule
        \textbf{Method} & \multicolumn{3}{c|}{\textbf{QG}} & \multicolumn{2}{c}{\textbf{VOG}} \\
        & \textbf{B-4 $\uparrow$} & \textbf{R-L $\uparrow$} & \textbf{MTR $\uparrow$} & \textbf{SSIM $\uparrow$} & \textbf{CLIP-T $\uparrow$} \\
        \midrule
        \rowcolor{red!10} 
        Qwen-VL & $\underline{\textbf{75.50}}$ & $\underline{\textbf{77.20}}$ & $\underline{\textbf{81.80}}$ & $\underline{\textbf{59.5}}$ & $\underline{\textbf{40.2}}$ \\
        
        LlaMA & $74.74$ & $76.51$ & $80.50$ & $57.5$ & $39.0$ \\
        \rowcolor{red!10} 
        LLaVA & $73.62$ & $75.13$ & $80.69$ & $57.7$ & $38.3$ \\

        InstructBLIP & $72.10$ & $75.61$ & $76.41$ & $55.9$ & $38.7$ \\
        \rowcolor{red!10} 
        mPLUG-Owl & $71.23$ & $72.66$ & $76.10$ & $56.4$ & $38.6$ \\
     
        VisualGLM & $58.63$ & $60.61$ & $64.06$ & $47.3$ & $33.1$ \\
        \bottomrule
    \end{tabular}
}
\caption{Detailed performance of \texttt{\textbf{CmOS}} with different base models. (MTR=METEOR) $\uparrow$: higher is better.}
\label{tab:basemode}
\end{table}

\section{Guideline of Human Evaluation}
Table \ref{gl} presents the evaluation form used to guide human annotators, consisting of three sections: case details, question evaluation, and visual option evaluation.
\begin{table*}[htbp]
\centering
\scriptsize
\resizebox{0.8\textwidth}{!}{
    \begin{tabular}{l|ccc|cc|cc|c}
        \toprule
        \textbf{Method} & \multicolumn{3}{c|}{\textbf{Subject}} & \multicolumn{2}{c|}{\textbf{Modality}} & \multicolumn{2}{c|}{\textbf{Grade}} & \textbf{AVG} \\
        \textbf{MTR}$\uparrow$ & \textbf{NAT} & \textbf{SOC} & \textbf{LAN} & \textbf{TXT} & \textbf{IMG} & \textbf{G1-6} & \textbf{G7-12} &  \\
        \midrule
        \rowcolor{white!10} 
        0-shot & $32.8$ & $29.2$ & $24.0$ & $29.2$ & $32.1$ & $31.2$ & $30.4$ & $30.8$ \\
        \rowcolor{red!10}
        1-shot & $46.9$ & $43.1$ & $33.7$ & $45.3$ & $46.2$ & $44.2$ & $48.7$ & $45.5$ \\
        3-shot & $51.2$ & $49.4$ & $42.3$ & $51.2$ & $50.3$ & $51.0$ & $50.4$ & $50.8$ \\
        \rowcolor{red!10}
        VL-T5 & $54.4$ & $48.1$ & $48.3$ & $54.5$ & $51.2$ & $52.8$ & $54.5$ & $53.1$ \\
        MultiQG-TI & $59.4$ & $48.7$ & $50.5$ & $57.2$ & $54.5$ & $56.8$ & $55.3$ & $56.0$ \\
        \rowcolor{red!10}
        Multimodal-CoT & $65.1$ & $57.9$ & $56.4$ & $62.1$ & $59.1$ & $60.7$ & $62.1$ & $61.4$ \\
        CoE & $72.3$ & $68.3$ & $63.5$ & $70.1$ & $67.8$ & $68.5$ & $70.0$ & $69.1$ \\
        \rowcolor{red!10}
        \texttt{\textbf{CmOS}} & $\underline{\textbf{80.9}}$ & $\underline{\textbf{84.1}}$ & $\underline{\textbf{72.6}}$ 
        & $\underline{\textbf{81.2}}$ & $\underline{\textbf{82.3}}$ & $\underline{\textbf{83.3}}$ & $\underline{\textbf{78.4}}$ 
        & $\underline{\textbf{81.8}}$ \\
        \hline
        \hspace{1em}w/o OQRM & $62.4$ & $64.9$ & $45.6$ & $62.4$ & $64.6$ & $65.4$ & $60.3$ & $63.5$ \\
        \hspace{1em}w/o Discriminator	&$71.7$	&$74.2$	&$63.6$	&$71.3$	&$72.5$	&$73.9$	&$69.7$	&$72.2$\\
        \bottomrule
    \end{tabular}}
\caption{Additional automatic evaluation results of question generation. (MTR=METEOR) $\uparrow$: higher is better.}
\label{mt}
\end{table*}

\begin{figure*}[t]
  \includegraphics[width=\linewidth] {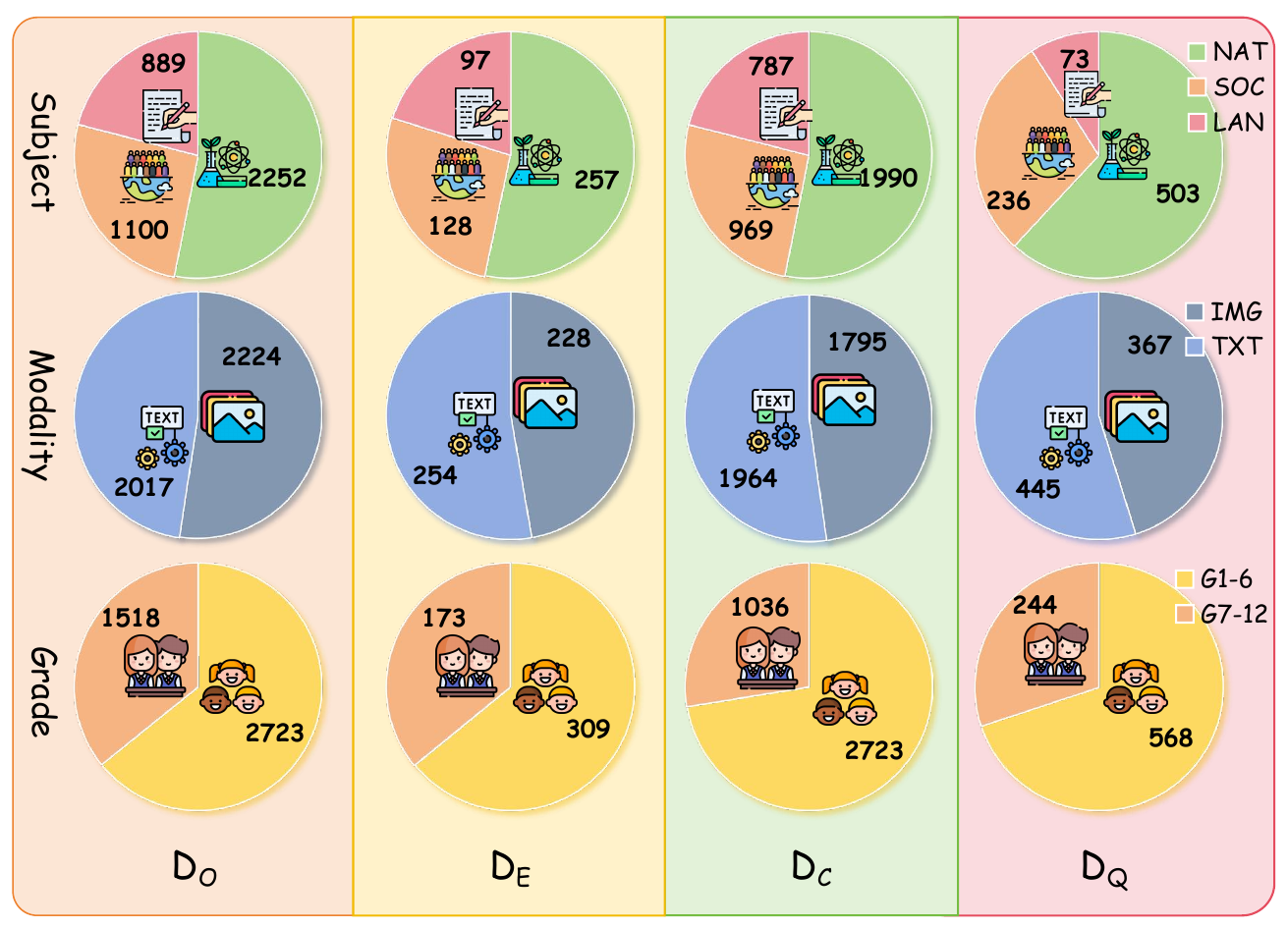}
  \caption{Dataset statistics of ScienceQA test benchmark and our test sets.  Question types: NAT = natural science, SOC = social  science, LAN = language science, TXT = containing  text context, IMG = containing image context, G1-6 = grades 1-6, G7-12 = grades 7-12.}
  \label{fig8}
\end{figure*}

\begin{figure*}[t]
\centering
  \includegraphics[scale=0.8]{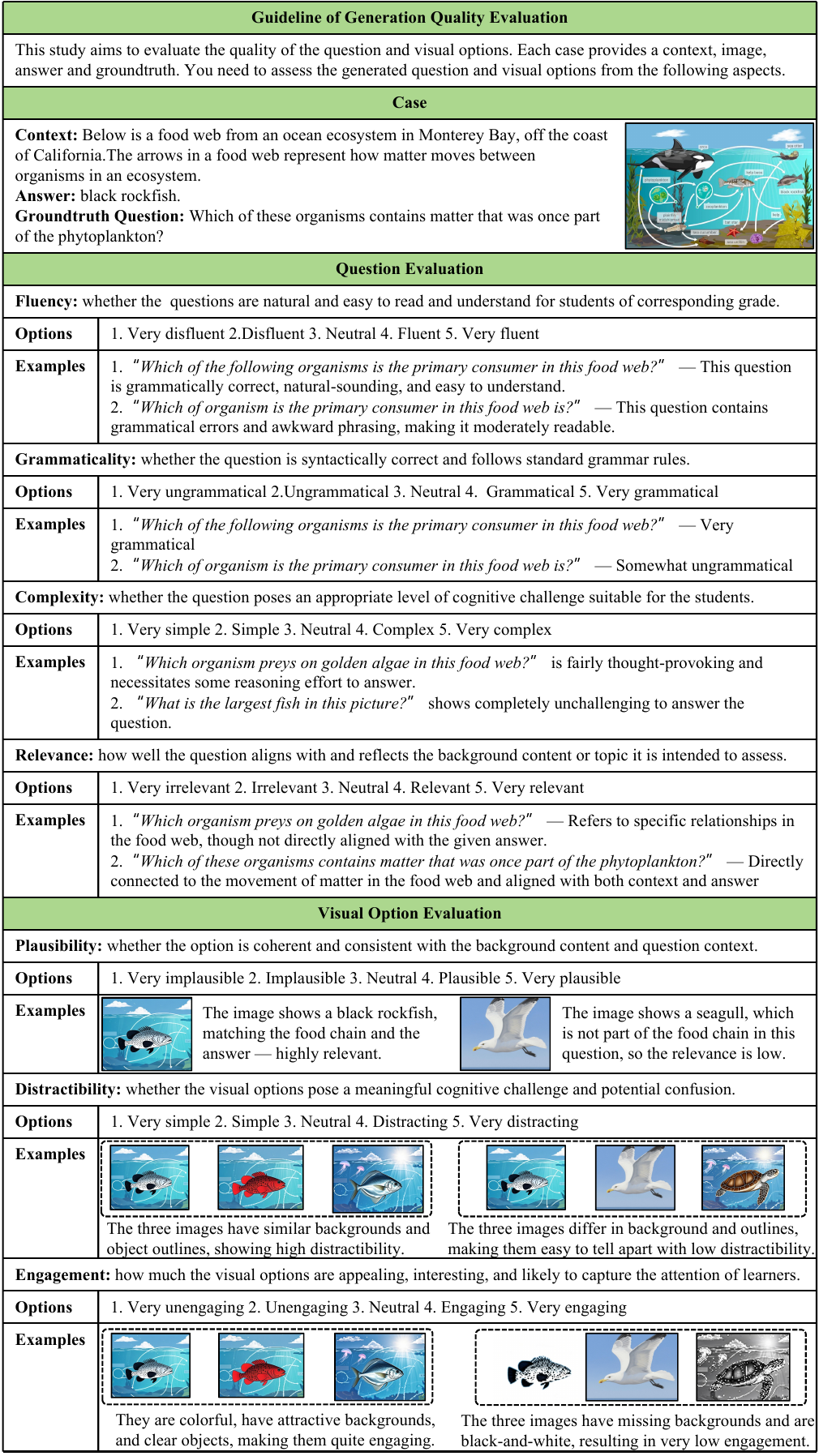}
  \caption{Guideline of human evaluation for question and visual option generation quality.}

\end{figure*}
\label{gl}

\end{document}